%% file: acl_latex.tex
\documentclass[11pt]{article}

\usepackage[]{acl}

\usepackage{times}
\usepackage{latexsym}
\usepackage{booktabs}
\usepackage{makecell}
\usepackage{multirow}
\usepackage{multicol}
\usepackage{CJKutf8}
\usepackage{graphicx}
\usepackage{algorithm}
\usepackage{algorithmic}
\usepackage{hyperref}
\usepackage[T1]{fontenc}
\usepackage{bbding}
\usepackage[utf8]{inputenc}

\usepackage{microtype}

\newcommand{\myparagraph}[1]{\vspace{2pt}\noindent{\bf{#1}}~~}

\newcommand{\sep}{[\textsc{sep}]}

\title{LiveChat: A Large-Scale Personalized Dialogue Dataset Automatically Constructed from Live Streaming}

\author{
   Jingsheng Gao\textsuperscript{1,2}\thanks{\; Work done during an internship at Xiaobing.AI},
   Yixin Lian\textsuperscript{2},
   Ziyi Zhou\textsuperscript{2},
   Yuzhuo Fu\textsuperscript{1}\thanks{\; Corresponding Author},
   Baoyuan Wang\textsuperscript{2}\footnotemark[\value{footnote}]\\
    \textsuperscript{\rm 1} School of SEIEE, Shanghai Jiao Tong University, China\\
	\textsuperscript{\rm 2} Xiaobing.AI \\
	{\tt\{gaojingsheng, yzfu\}@sjtu.edu.cn } \\
        {\tt\{lianyixin, zhouziyi, wangbaoyuan\}@xiaobing.ai }
}
\begin{document}
\begin{CJK*}{UTF8}{gbsn}
\maketitle

\input{sections/0abstract}

\input{sections/1introduction}

\input{sections/2relatedwork}

\input{sections/3datasetconstruction}
\input{sections/4models}

\input{sections/5experiments}

\input{sections/6conclusion}

\input{sections/7limitations}

\input{sections/8ethicalconsideration}

\input{sections/10Acknowledgements}

\bibliography{anthology,custom}
\bibliographystyle{acl_natbib}

\appendix

\input{sections/9appendix}

\end{CJK*}
\end{document}

%% file: sections/0abstract.tex

\begin{abstract}
Open-domain dialogue systems have made promising progress in recent years. While the state-of-the-art dialogue agents are built upon large-scale text-based social media data and large pre-trained models, there is no guarantee these agents could also perform well in fast-growing scenarios, such as live streaming, due to the bounded transferability of pre-trained models and biased distributions of public datasets from Reddit and Weibo, etc. To improve the essential capability of responding and establish a benchmark in the live open-domain scenario, we introduce the LiveChat dataset, composed of 1.33 million real-life Chinese dialogues with almost 3800 average sessions across 351 personas and fine-grained profiles for each persona. LiveChat is automatically constructed by processing numerous live videos on the Internet and naturally falls within the scope of multi-party conversations, where the issues of Who says What to Whom should be considered. Therefore, we target two critical tasks of response modeling and addressee recognition and propose retrieval-based baselines grounded on advanced techniques. Experimental results have validated the positive effects of leveraging persona profiles and larger average sessions per persona. In addition, we also benchmark the transferability of advanced generation-based models on LiveChat and pose some future directions for current challenges. \footnote{The code and dataset will be publicly available at \href{https://github.com/gaojingsheng/LiveChat}{https://github.com/gaojingsheng/LiveChat}.}
\end{abstract}

%% file: sections/1introduction.tex
\section{Introduction}

\begin{figure}[htbp]
	\centering
	\includegraphics[width=0.44\textwidth]{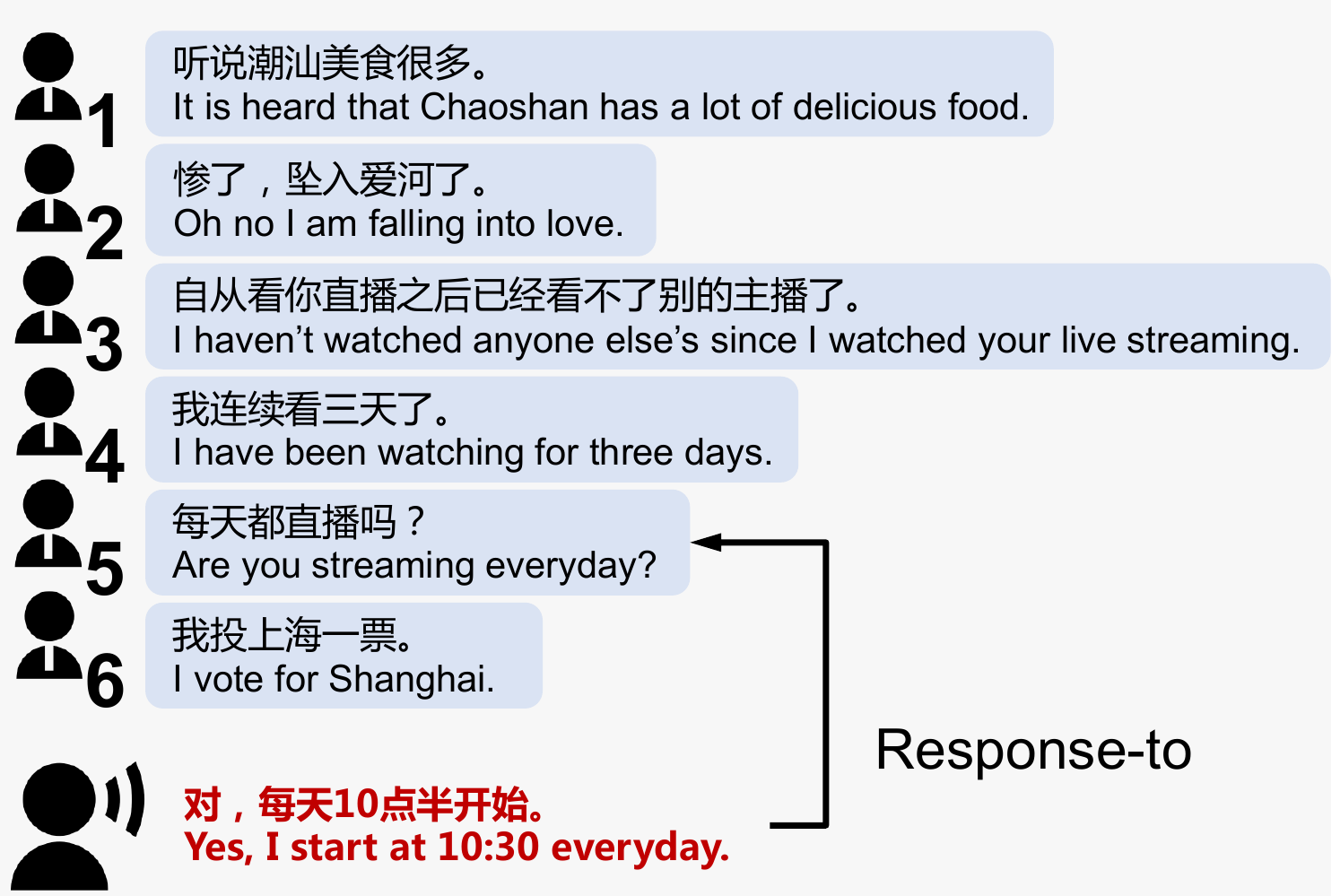}
	\caption{A session example of LiveChat. A streamer will respond to one audience's comment from the comments area.}
	\label{fig:example}
\end{figure}

Building dialogue systems to converse naturally with humans has been one of the longest-running goals in artificial intelligence ~\citep{xiaoice, roller-etal-2021-recipes}.  To usher that chatbot response properly in diverse scenarios, it is desirable to train a conversational agent based on massive large-scale datasets with multiple domains. Current dialogue datasets mainly leverage online forum posts to build reply-to relationships between users, such as Reddit ~\citep{mazare-etal-2018-training, zhong-etal-2020-towards} and Weibo ~\citep{weibo, pchatbot}. Despite the scalability and diversity of current dialogue corpora, dialogue models pre-trained on these conversation datasets can not perform effectively when applied to a completely new domain, such as live streaming. The reason lies in the intrinsic domain gap between online-post constructed data and those required in downstream conversational tasks. Even recent state-of-the-art (SOTA) dialogue models built upon large pre-trained language models (PLMs) like LaMDA ~\citep{Thoppilan2022LaMDALM} and ChatGPT\footnote{\href{https://openai.com/blog/chatgpt}{https://openai.com/blog/chatgpt}} heavily rely on publicly available text-only data.
These large pre-trained models' distributions remain different across domains ~\citep{Zeng2022SocraticMC} and are distinct from those of models learning the information contained in other modalities, video as an example. 
%
\begin{table*}
   
    \centering
    \footnotesize
    \setlength{\tabcolsep}{1.5mm}
     \resizebox{0.95\textwidth}{!}{%
	\begin{tabular}{ccccccc}
	\toprule
        Dataset & Data Source & Dialogues & Persona & Addressee & Avg. Sessions & Language \\
        \midrule
        PersonaChat~\citep{personachat} & Crowdsourced & 10,907  &  \CheckmarkBold  & \XSolidBrush & 8.69 & English \\
        PCR~\citep{mazare-etal-2018-training} & Online Posts & 700,000,000 &  \CheckmarkBold  & \XSolidBrush & 53.0 & English \\
        PersonalDialog~\citep{weibo} & Online Posts & 20,830,000 & \CheckmarkBold & \XSolidBrush  & 6.64 & Chinese  \\
        PEC~\citep{zhong-etal-2020-towards} & Online Posts & 355,000 &  \CheckmarkBold  & \XSolidBrush & 26.0 & English \\
        PchatBot~\citep{pchatbot} & Online Posts & 198,875,796 & \CheckmarkBold & \XSolidBrush & 7.58 & Chinese \\
        MSC~\citep{xu-etal-2022-beyond} & Crowdsourced & 5,001  &  \CheckmarkBold  & \XSolidBrush & 42.9 & English \\
        DuLemon~\citep{xu-etal-2022-long} & Crowdsourced & 27,501  &  \CheckmarkBold  & \XSolidBrush & 16.3 & Chinese \\
        

        
        Linux-IRC ~\citep{elsner-charniak-2008-talking} & Online Chatroom &  2,500 &  \XSolidBrush & \CheckmarkBold & - & English \\
        
        Ubuntu-IRC ~\citep{UbuntuIRC} & Online Chatroom & 77,563 &  \XSolidBrush & \CheckmarkBold & - & English \\

        INTERVIEW~\citep{interview} & Interview Transcripts & 105,000  &  \XSolidBrush & \XSolidBrush & - & English \\  
        RealMedDial$^*$~\citep{xu-etal-2022-realmeddial} & Short Videos & 2,637 &  \CheckmarkBold & \XSolidBrush & 44.7 & Chinese \\  
        \midrule
		LiveChat (ours) & Live Videos & 1,332,073 & \CheckmarkBold & \CheckmarkBold & 3795 & Chinese \\
		
		\bottomrule
	\end{tabular}
 
    }
	\caption{\label{tab:dataset} Comparison between our dataset and other existing open-domain dialogue datasets (mainly for tasks of personalized dialogue generation and addressee recognition). $^*$ for the medical domain. Persona represents whether there are personal profiles in the dataset. Addressee means if the dataset contains reply-to labels for addressee recognition problem in MPCs. Avg. Sessions denotes the average session number per persona and - means it is not mentioned in the dataset. Note that LiveChat can automatically and continuously construct dialogue sessions from videos while other video-sourced works like RealMedDial depend on crowdworkers.}
\end{table*}

Video is also an important dialogue data source in the wild with great diversity. As a form of popular video-based conversations, streaming is a broadcasting scenario that transcribes and broadcasts at the same time, which involves entertainment, life-sharing, education and so on~\citep{LiveCommerence}. Such video-based conversations are one of the main ways human beings spread and exchange information efficiently in their daily lives and are naturally in line with the way people communicate. They are also the desired sources of dialogue datasets that are vitally significant in training large-scale dialogue models for homologous downstream virtual human scenarios, such as Virtual YouTubers, Virtual Employees, and Virtual Celebrities. Nevertheless,
works that extract data from online videos do not receive enough attention although video-sourced dialogues are more life-oriented and naturally abundant. 

Current video-sourced spoken corpora can be separated into two main categories ~\citep{mahajan-shaikh-2021-need}: scripted and unscripted. The former refers to planned dialogues such as movie and TV scripts ~\citep{CornellMovie, li-etal-2016-persona}. The latter means spontaneous conversations in real situations, for instance, the interview dataset of ~\citet{interview}.
However, these previous video-sourced dialogues can not meet the scale of training a satisfied chatbot, owing to the trouble of continuously obtaining and processing various kinds of videos, and troubles of extracting valid dialogue sessions from them. For example, it is challenging to build valid dialogue sessions automatically from movies without human annotators. 
Thus, a large-scale video-sourced dialogue dataset in live streaming is essential for facilitating research in this area. The live broadcast is a typical one-to-many chat scene, which generally involves one streamer and multiple audiences. The challenge of building such a dataset lies in retrieving the reply-to relationships between the streamers and audiences. Unlike post-based social media with clear links between posts and replies, the streamer's responses in the live scene have no explicit reply-to relationships with audiences' comments. 

To tackle the aforementioned problems, in this paper, we propose a novel and automatic video-sourced dialogue-constructing method and build a large-scale personalized dialogue dataset from the live streaming domain, named \textbf{LiveChat}. It is a non-trivial work since this dataset originates from a video-based source, distinct from most previous text-sourced data. Meanwhile, as far as we know, this is almost the only work that can effectively and endlessly extract dialogue sessions from videos.



As illustrated in ~\citet{huang2020challenges}, one of the main challenges of existing open-domain chatbots is lacking a consistent personality as these agents are trained over different dialogues each with no or limited speaker information, while LiveChat naturally contains distinctive persona features (especially for streamers). To promote research in this field, we collect publicly available information for each streamer and add manual annotations to create the persona profiles, with individual information anonymized for privacy concerns. Compared to the previous personalized dialogue datasets ~\citep{personachat, mazare-etal-2018-training, weibo, zhong-etal-2020-towards, pchatbot, xu-etal-2022-long}, our dataset provides more fine-grained persona profiles, and more importantly, the average session number of each speaker exceeds previous ones extraordinarily, as shown in Table \ref{tab:dataset}. This proves to be beneficial for personalized dialogue modeling.




Moreover, live streaming is also a multi-party conversation (MPC) scene involving more than two interlocutors. An example of LiveChat is illustrated in Figure \ref{fig:example}. During the streaming process, a streamer naturally has to recognize which audience to reply to. We collect public live videos and process the streamer's responses and all audiences' comments to form multiple sessions of dialogues where each session contains a streamer's response and multiple candidates of addressee comments. A reply-to-whom matching method is brought forward to accurately find the correct candidate for a streamer's response. In this way, we can leverage the reply-to-whom relationship to build datasets for two classical tasks: response modeling and addressee recognition. Our proposed two classical dialogue tasks in LiveChat can help solve the MPC problem in a unified dataset, essential for building a practical dialogue agent in live streaming.

To sum up, our main contributions are as follows:

\begin{itemize}
    \item We propose a large-scale personalized dialogue dataset LiveChat with a unique automatic dialogue-constructing method for countless live streams in the wild. To the best of our knowledge, our LiveChat is not only the largest video-sourced dialogue dataset, which contains detailed persona profiles and the largest average sessions per persona, but also the largest MPC dataset for addressee recognition released to the community.

    
    \item Sufficient experiments on two benchmark tasks: Response Modeling and Addressee Recognition, prove that our persona selection method is beneficial and larger average sessions per persona do help the modeling of the dialogue. We design retrieval baselines with considerable performance on both tasks to facilitate further research and build more genuine live-domain dialogue systems. 
    
    
    \item We further investigate transfer learning of generation models and illustrate that pre-trained dialogue models perform poorly under the video-sourced data after fine-tuning, while large PLMs exhibit richer informativeness but worse relevance under few-shot settings. This arouses the interest in exploring domain adaptation with large PLMs in such video-sourced datasets.
    
    
\end{itemize}

%% file: sections/2relatedwork.tex
\section{Related Work}

\myparagraph{Dialogue Datasets} A qualified open-domain dialogue model is usually trained on sufficient supervised datasets. Due to the accessibility and characteristics of social media, the current large-scale open-domain dialogue datasets are mainly constructed from text-based social media, such as Reddit~\citep{mazare-etal-2018-training, zhong-etal-2020-towards}, Douban~\citep{DouBan}, and Weibo~\citep{pchatbot}. 
Besides, a large-scale dataset with persona annotations is essential in building a personalized dialogue system. The persona profiles utilized in current persona datasets can be generally classified into two categories: basic profiles and text profiles. The basic profiles in ~\citet{weibo} and ~\citet{pchatbot} are composed of personality traits like age, gender, and location. The text profiles are mainly composed of crowdsourced ~\citep{personachat, xu-etal-2022-long} or automatically collected~\citep{mazare-etal-2018-training, zhong-etal-2020-towards} descriptive persona sentences. In LiveChat, we collect more fine-grained basic profiles and text profiles, with extraordinarily larger average sessions per persona than in previous works.

Furthermore, multi-party dialogue datasets are crucial when occurring conversations consisting of more than two speakers. However, most existing MPC datasets~\citep{CornellMovie, UbuntuIRold, MEISD} have no explicit reply-to-whom annotations, and thus can not be leveraged in addressee recognition.  ~\citet{elsner-charniak-2008-talking} manually group sentences of disentangled conversations into separated sessions in \texttt{Linux} IRC. ~\citet{UbuntuIRC} propose a larger MPC dataset manually annotated with reply-to structure from the \texttt{Ubuntu} IRC channel, which extremely prompts the research in MPC problems. Our LiveChat naturally originates from a multi-party scenario, whose size also remarkably exceeds previous ones, credit to the automatically reply-to-whom matching method.

As for those spoken dialogue corpora ~\citep{xu-etal-2022-realmeddial, interview, li-etal-2016-persona, CornellMovie}, most are pre-scripted or manually transcribed, intrinsically difficult to scale up because of the restricted video- or audio-based sources where people can effortlessly extract valid dialogue sessions. 



\begin{figure*}[htbp]
	\centering
        \includegraphics[width=0.95\textwidth]{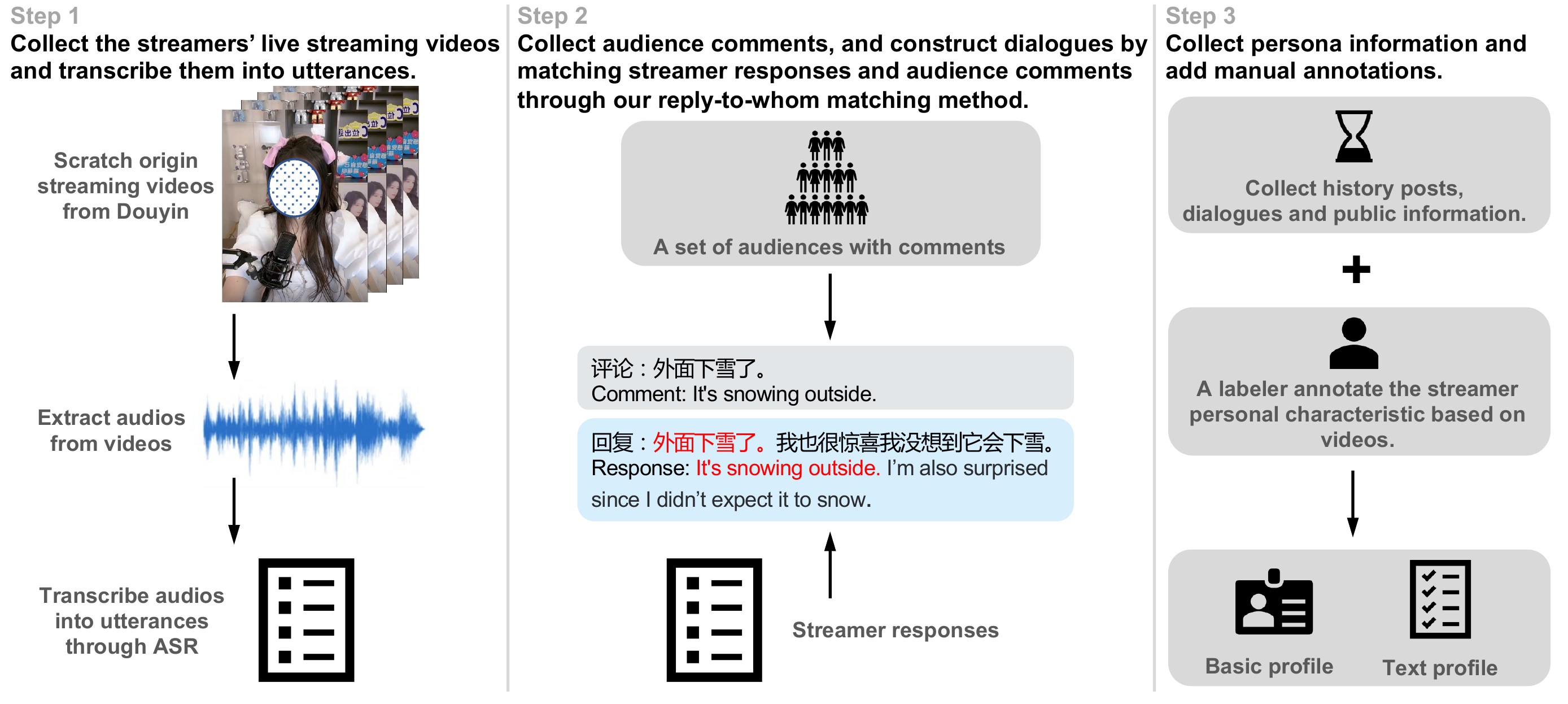}
	\caption{The whole construction process of LiveChat.}
	\label{fig:DataConstruction}
\end{figure*}

\myparagraph{Personalized Response Modeling}
Early works use  explicit persona profiles from predeﬁned information or implicit persona vectors from dialogue history to generate personality-coherent responses.
Explicit models use persona descriptions, attributes, or extracted profiles to learn personalized response modeling. ~\citet{kim2014acquisition} leverages a persona knowledge base to extract predefined triples and entities in a retrieval-based dialogue system. 
~\citet{ijcai2018p595} propose an explicit persona model to generate personalized responses based on a pre-specified user profile. ~\citet{ijcai2019p0721} propose a memory-augmented architecture to exploit persona information from context to generate diverse and sustainable conversations. On the other hand, implicit methods like ~\citet{zhang2019consistent} generate consistent responses by maintaining certain features related to topics and personas, while ~\citet{li2021dialogue} encodes all the dialogue history of a speaker into the implicit persona. ~\citet{Lessismore} design a personality selecting module to obtain abundant and accurate persona information from the user dialogue history. In LiveChat, we leverage explicit persona information to maintain persona consistency. 


 

\myparagraph{Addressee Recognition}
Addressee recognition which is also named explicit addressee modeling aims at understanding who speaks to whom in a multi-party conversation.
Previous works mainly focus on predicting the targeting addressee of the last utterance in one conversation ~\citep{ouchi-tsuboi-2016-addressee, zhang2018addressee}. Later on, a who-to-whom model for predicting all the missing addressees to understand the whole conversation was introduced by ~\citet{LeAddressee}. ~\citet{gu-etal-2021-mpc} further leverages a pre-trained language model for learning this problem in a unified manner.
We follow this learning paradigm, and furthermore, are able to investigate personalized addressee recognition in LiveChat attributed to the available persona profiles.




%% file: sections/3datasetconstruction.tex
\section{Dataset Construction}
\label{sec:datasetconstruction}
\begin{algorithm}[tb]
    \caption{Dialogue construction through reply-to-whom matching method.}
    \label{alg:algorithm}
    \textbf{Input}: The streamer responses $\mathcal{R}$ and audience comments $\mathcal{C}$; each sentence is accompanied with timestamp $T$; max response time interval $\Delta  t$; length ratio threshold $\tau$; matching function $\mathcal{F}$.  \\
    \textbf{Output}: Matched dialogues $\mathcal{D}$. 
    
    \begin{algorithmic}[1] 
        \STATE \textbf{Step 1: } $c_i \gets \mathcal{C} $ \quad $\rhd$ \textit{Traverse all comments}
        \STATE $r_j \gets \mathcal{R}$ where $0 \le T_{rj} - T_{ci} \le \Delta t$ \quad $\rhd$ \textit{Traverse the responses during time interval}
        \STATE $c_i \to \mathcal{M}_j$ if $\mathcal{F}(c_i,r_j)=1$ \quad  $\rhd$ \textit{Record all matched comments of response j in a set $\mathcal{M}_j$}
        \STATE \textbf{Step 2: } $r_m \gets R$ \quad $\rhd$ \textit{Traverse all responses}
        \STATE \textbf{if} {$\mathcal{M}_m \ne \oslash$,  $c_n \gets \mathcal{M}_m$} \textbf{then} $\rhd$  \textit{Traverse matched comments in reverse order.}
        \STATE \quad \textbf{if} {$r_m[-1] = .$ or $?$ \textbf{then} \quad $\rhd$ \textit{Detect if the response with an ending punctuation}} 
        \STATE \quad \quad \textbf{if} {$\frac{len(r_m)}{len(c_n)}  > \tau $} \textbf{then}
        \STATE \quad \quad \quad$(c_n,r_m) \to \mathcal{D}$, break \quad $\rhd$ \textit{Add matched dialogue pairs.}
        \STATE \quad \quad \textbf{else} $r_m \to r_{m+1}$ \quad $\rhd$ \textit{Merge current response sentence into next one}
        \STATE \quad \textbf{else} $r_m \to r_{m+1}$ \quad $\rhd$ \textit{Merge current response sentence into next one}
        \STATE \textbf{return} $\mathcal{D}$
    \end{algorithmic}
\end{algorithm}

\subsection{Dataset Overview}
The raw data constructed in LiveChat are collected from Douyin\footnote{\href{https://www.douyin.com}{https://www.douyin.com}} (Chinese Tiktok), one of the largest Chinese live streaming and short video platform with over 10 million streamers and around 800 million users. We selected 351 representative streamers that interact and chat with the audiences frequently. By capturing the publicly available streamers' live videos and the audiences' comments in the broadcast room for a long time, we retrieved massive video clips with a huge amount of comments. 



The whole dialogue construction process is shown in Figure \ref{fig:DataConstruction}, consisting of three steps. The first two steps are to construct dialogue sessions by processing videos and matching audience comments with streamer responses, and the last step is to enrich the dataset with fine-grained persona profiles, including basic profiles and text profiles.

\subsection{Dialogue Construction}

Firstly we  have to collect the raw spoken texts of the streamers. Since the original data are in the form of video clips, we need to transcribe them into text utterances. A video format converter is utilized to extract the voice content. Then we leverage an automatic speech recognition (ASR) model\footnote{\href{https://www.volcengine.com}{https://www.volcengine.com}} to transcribe these voice clips into texts with timestamps, and this model is fine-tuned on a large-scale pan-entertainment dataset. Consequently, the raw data is transcribed into the streamer's spoken texts. Details of ASR are  illustrated in Appendix \ref{sec:data const}.

Secondly, we collect the raw audience comments and propose a reply-to-whom matching method to retrieve the reply-to relationships between streamers and audiences. Our proposed matching method is mainly based on the observations particularly apt to the streaming scenario: the streamer will reply to one audience in the comments area after that audience sent the message for a while. And usually, the streamer will repeat or summarize the audience's comment before responding to it, which helps the rest of the audiences understand what the streamer is talking about. We simply focus on extracting valid dialogue sessions based on the above observations and filter out others that are not satisfied. 
On this basis, the pseudocode of the whole matching process is illustrated in Algorithm \ref{alg:algorithm}. For each audience comment, we go through all the transcribed spoken utterances by the streamer within one minute. If there exists a repetition or summarization of this comment in the transcribed streamer's utterance,
they will be recorded as a matched pair.
Note that we apply a combination of BOW (bag of words) and pre-trained Chinese BERT~\citep{ChineseBERT} as the matching function.
After retrieving the matched pairs, we iteratively concatenate the transcribed streamer's utterances to meet the ending punctuation and satisfy the required threshold $\tau$ 
 for sufficient length, because the transcribed response from the ASR tool can sometimes be a broken sentence from what the streamer originally expresses.
In addition, if a response matches several comments, we choose the closest one in time.

For each constructed dialogue pair, the response will repeat the comment. To prevent models from overfitting in this kind of manner, we remove the repetition prefix of each response.
Besides, considering the specificity of this scenario, we filter out noisy pairs such as
"\texttt{谢谢 **(Thanks to **)}" or "\texttt{欢迎 **(Welcome **)}" which miss valuable dialogue information.
Finally, we can construct the dataset based on such matched pairs. 

\subsection{Persona Extraction}

The last step is to construct detailed persona profiles in LiveChat, which are composed of basic profiles and text profiles. Following the work of PersonalDialog ~\citep{weibo} and Pchatbot ~\citep{pchatbot}, the basic profiles contain age, gender, and location. Except these, the basic profile in LiveChat also includes streamer characters and live room information such as live time, fans number, live streaming style, and so on.
Part of this information can be retrieved from the live room or the streamers' homepages, besides, we crowdsource a set of questions and each annotator is required to label those missing contents by watching these streamers' streaming videos. Details about data privacy and annotators are elaborated in Ethical Consideration and Appendix \ref{sec:data const}. 

The text profile is composed of several sentences which describe the streamer's personal habits or characteristics. Sentences in the text profile are extracted in two ways: rules-based and classifier-based. Similar to ~\citet{mazare-etal-2018-training} and ~\citet{zhong-etal-2020-towards}, we collect persona sentences from all history spoken utterances and posts the streamer spoke or wrote on Douyin by rules. The final selected sentences must satisfy the following requirements: 1) between 4 and 20 words; 2) the contents include "\texttt{我(I)}"; 3) at least one verb; 4) at least one noun or adjective. Besides this, we train an additional persona classifier to further refine the text profiles. In detail, the classifier-based method means to discriminate if a single sentence contains persona facts by a learned classifier, which in our case is trained from DuLemon ~\citep{xu-etal-2022-long}. 

\subsection{LiveChat}

We combine each pair of audience comments and streamer responses along with each streamer's corresponding persona to create LiveChat, the first large-scale personalized dialogue dataset from the live streaming domain. It is worth noting that each session in LiveChat contains not only the pairs of comments and responses but also several comments candidates within the same period, details illustrated in the appendix \ref{sec:data const}. Although the LiveChat we discussed in this paper consists of single-turn-only dialogues, the multi-turn dialogues can be easily built by continuously tracing the interaction between the streamer and the same audience in a range of time. Data privacy in LiveChat including persona profiles is assured by carrying out the transformation, deletion, and anonymization of personal information as illustrated in Ethical Consideration. 

With LiveChat, we propose that two benchmark tasks should be considered: (1) Response Modeling; (2) Addressee Recognition. The matched dialogue pairs can be directly leveraged in response modeling, while the other candidates of comments can be grouped together for training the addressee recognition task.

%% file: sections/4models.tex
\section{Models}

\subsection{Task Definition}
\label{sec:tasks}

\myparagraph{Response Modeling} Suppose we have a dialogue dataset $\mathcal{D}=\left\{\left(C_i, R_i, P_i\right)\right\}_{i=1}^n$, where $\forall i \in {1,...,n}$, $C_i$ is the input dialogue context, $R_i$ is the response, and $P_i$ is the corresponding persona profile for the respondent of $C_i$. The goal is to learn a dialogue model $g$ from $\mathcal{D}$, where for any new input context $C_j$, $g$ can generate a response $R_j$ based on its given persona $P_j$. 

Previous works chiefly include retrieval-based and generation-based methods.
To study the quantitative influence of our proposed persona profiles, we apply the retrieval-based architecture for the main experiments.
As for the study of the transferable performance of advanced models in LiveChat, most generation-based ones are investigated.

\myparagraph{Addressee Recognition}
Given a streamer $S_i$ with persona profile $P_i$, a response $R_i$, and a set of comments $C_{i1}, C_{i2}, ..., C_{im}$, where $\forall j \in {1,...,m}$, each comment $C_{ij}$ is associated with an audience $A_j$. The goal is to recognize which $C_{ij}$ (or $A_j$) the $R_i$ targets.
Note that the purpose of this task is to identify the appropriate addressee comment instead of the appropriate streamer reply in response modeling.
Dataset details about the settings of candidate comments can be seen in Appendix \ref{appendix:Selection}.

\subsection{Architecture}

To investigate how existing dialogue baseline models can be leveraged in LiveChat, we build three retrieval-based models for response modeling and addressee recognition. Besides, five generation-based pre-trained language models (PLMs) are taken into account to study transfer learning on LiveChat. Details of our utilized models in this paper are described below. 

\label{sec:models}
\subsubsection{Retrieval-based models}
\begin{figure}[htbp]
	\centering
	\includegraphics[width=0.44\textwidth]{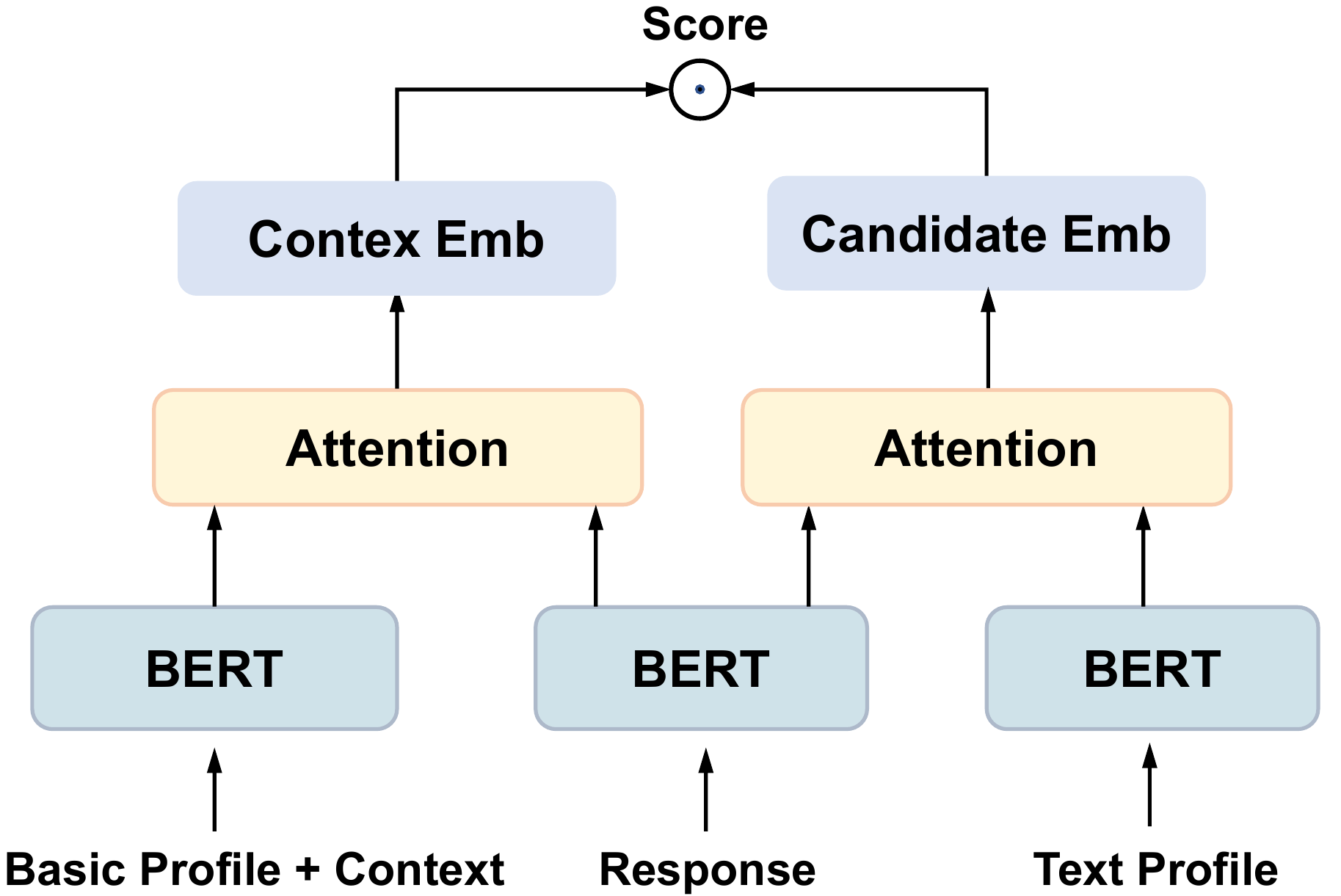}
	\caption{Our retrieval-based architecture.}
	\label{fig:model}
\end{figure}

\myparagraph{CoBERT} The overall architecture of our retrieval-based persona model is depicted in Figure \ref{fig:model}, which is inspired by \citet{zhong-etal-2020-towards}. 

We encode context, response, and text profile by separated BERT~\citep{devlin-etal-2019-bert}. Given an input user context, we leverage the basic profile as the streamer's initialized embedding, and a $\sep$ token is added between the basic profile and context. During our experiments, we only use the streamer ID information instead of all annotations. As for the multiple text profile sentences, we concatenate them with [SEP] to meet the length of maximum input tokens. After retrieving three individual representations, two co-attention modules~\citep{zhong-etal-2020-towards} are implemented for better feature fusion. Finally, we obtain context embedding and candidate response embedding, then apply dot product to compute the matching score and calculate cross-entropy loss to optimize the full network.

\myparagraph{TwinBERT} Current advanced retrieval-based models can be generally classified into context-response matching double-stream frameworks~\citep{humeau2019poly, lu2020twinbert} and PLMs-based single-stream frameworks~\citep{DBLP:conf/cikm/GuLLLSWZ20}. To keep the bi-encoder model consistent with CoBERT, we also adopt the attention module into TwinBERT~\citep{lu2020twinbert}, but without extra inputs of persona profiles to compare the effects of personal information. 

\myparagraph{BERT} BERT ~\citep{devlin-etal-2019-bert} is a typical single-stream network. The interaction and aggregation operations can be performed in a unified way by feeding the concatenation of the context and the response candidate into the model. During the inference stage, we can sort the output scores between the context and all response candidates to finally obtain the matched response. Note that in experiments of CoBERT, TwinBERT, and BERT, we use the pre-trained BERT checkpoint of the Chinese version.

\subsubsection{Generation-based models}

\myparagraph{BART}~\citep{CPT} is a denoising autoencoder for pre-training sequence-to-sequence model and pre-trained by reconstructing the original text from the arbitrary corrupting text, which has been a universal transformer-based baseline PLM. 

\myparagraph{CDialGPT}~\citet{CDialoGPT} proposed a Chinese GPT pre-trained from a large version of the open-domain dialogue dataset. The dataset sources originate from Chinese online forums, such as Weibo and Douban. 

\myparagraph{EVA2.0} is an encoder-decoder PLM for open-domain dialogue modeling ~\citep{EVA2}, whose architecture is similar to BART. This model is pre-trained on a 60GB high-quality dialogue dataset, which is composed of WDC-Dialogue ~\citep{EVA1} and some extra copra, like movie scripts or crowdsourcing datasets. WDC-Dialogue is sourced from Chinese social media and is the main training dataset of EVA2.0.

\myparagraph{GLM}~\citep{du-etal-2022-glm} is a large-scale model based on autoregressive blank infilling to unify all language tasks. The original Chinese GLM owns 10 billion parameters pre-trained on a Chinese corpus. 

\myparagraph{GPT3}~\citep{NEURIPS2020_1457c0d6} is an autoregressive language model with 175 billion parameters, which has shown engaging performance on many NLP tasks and exhibits powerful abilities in multilingual zero-shot, one-shot, and few-shot settings.

%% file: sections/5experiments.tex
\section{Experiments}

We train retrieval baselines for two tasks as described in Section \ref{sec:tasks}: response modeling and addressee recognition. We also investigate transfer learning of current popular generation-based models on LiveChat. Experimental settings including training details and evaluation metrics can be found in Section \ref{sec:train details}.




\subsection{Results of Response Modeling}

In this session, we fully investigate the influence of our persona profiles, the extraction methods for text profiles, and the impact of larger average sessions per persona. The main architecture follows the work of CoBERT \citep{zhong-etal-2020-towards}. Note that CoBERT without extra persona profile input is equal to TwinBERT \citep{lu2020twinbert}.


\myparagraph{Impact of Personas} The test performance of retrieval-based response modeling is shown in Table \ref{tab:retrival}. Obviously, CoBERT with text profile and basic profile achieves the best performance in our experimental settings, indicating both text profile and basic profile will facilitate the modeling of response. We attribute this to the fact that the basic profile is significant in denoting the corresponding speaker, and the text profiles include detailed personal descriptions which may have correlations with the candidate responses. An exclusive text profile achieves a higher score than a single basic profile, that is, detailed persona features of text profiles retrieve a more essential influence on model performance. 

\begin{table}
	\centering
	\footnotesize
	\setlength{\tabcolsep}{1.5mm}
	\begin{tabular}{l|cccc}
	\toprule
		
       Model & Recall@1 & Recall@2  & MRR  \\ 
        \midrule
        
         CoBERT & 68.72 & 75.58 & 76.25   \\
        
        \quad \textit{+ text profile} & 70.04 & 77.43  & 77.66  \\
        
         \quad \textit{+ basic profile} & 69.43 & 76.58  & 77.06  \\
        
        \quad \textit{+ text \& basic profile} & \textbf{72.18} & \textbf{79.58}  & \textbf{79.63} &  \\
        

	\bottomrule
	\end{tabular}
	\caption{\label{tab:retrival} Comparison of automatic evaluation metric results (\%) among different retrieval-based settings.}
\end{table}

\myparagraph{Impact of Average Sessions}
To study the influence of the length of average sessions per persona on the model performance, we conduct experiments on different settings of data scales and the number of persona IDs based on CoBERT along with complete persona profiles. Since the data scale is equal to the persona ID number times the average session number by person, and the same number of persona IDs with larger data scales and the same data scales with fewer IDs both indicate that there are more average sessions per persona. To reduce the influence of different scales of training data and make a fair comparison, we also keep the same data scale (100k) while decreasing the number of IDs from 150 to 15 as shown in Table \ref{tab:datascale}. We make sure the persona IDs of the test set are all seen before. Consequently, all of our testing persona IDs are incorporated into the training settings. 

Experimental results demonstrate: (1) Obviously, more average sessions with the same number of IDs will enhance the model to capture the speaker's personalized response. 
(2) The average number of sessions is more significant than the number of IDs for response modeling.
The priority of the number of sessions per persona also proves the superiority of our proposed dataset to other existing ones since LiveChat exceeds others extraordinarily in this indicator.

\begin{table} 
	\centering
	\footnotesize
	\setlength{\tabcolsep}{1.5mm}
	\begin{tabular}{cc|ccc}
	\toprule
		
        Data Scale & ID Num & Recall@1 & Recall@2 & MRR  \\
        \midrule
        
       400k & 150 & \textbf{69.39} & \textbf{77.87}  & \textbf{77.67}   \\
  
       100k & 150 & 67.86 & 74.99 &  75.63 \\

       100k & 50 & 67.65 & 75.95 &  75.95  \\

       100k & 15 & 68.78 & 77.25 &  77.09 \\

       40k & 150 & 64.01 & 71.57 & 72.50   \\

	\bottomrule
	\end{tabular}
	\caption{\label{tab:datascale} Test performance (in \%) under different data scales and number of persona IDs.}
\end{table}
		
        
       



        

\begin{table}
	\centering
	\footnotesize
	\setlength{\tabcolsep}{1.5mm}
	\begin{tabular}{cc|ccc}
		\toprule
		
       Persona Selection & Length & Recall@1  & MRR  \\
        \midrule
        
        -  &  0 &  69.43  & 77.06 \\
        
       
       rules + classifier & 256 & 71.09 & 78.49 \\

       random from user & 512 &  69.49 & 77.27 \\
   
        
       random from dataset & 512 &  69.46 & 76.92 \\
       
       rules & 512 &  71.07 & 78.55 \\
       
       
       classifier & 512 &  71.19 &  78.61  \\

       rules + classifier & 512 & \textbf{72.18} & \textbf{79.63}    \\
        
		\bottomrule
	\end{tabular}
	\caption{\label{tab:selection} Test performance (in \%) among different persona selection methods.}
\end{table}

\begin{table}
	\centering
	\footnotesize
	\setlength{\tabcolsep}{1.5mm}
	\begin{tabular}{l|cccc}
	\toprule
         Model & Recall@1 & Recall@2  & MRR  \\ 
        \midrule
        
        BERT & 62.29 & 75.38 & 74.59 \\ 

        TwinBERT & 58.76 & 72.52 & 71.92  \\
  
        CoBERT & 59.27 & 73.04 & 72.43\\

		\bottomrule
	\end{tabular}
	\caption{\label{tab:addressee}  Test performance (in \%) among different addressee recognition models.}
\end{table}


\begin{table*}
	\centering
	\footnotesize
	\setlength{\tabcolsep}{1.5mm}
        \resizebox{0.95\textwidth}{!}{%
	\begin{tabular}{lcc|cccc|cccc}
	\toprule
        & Pre-trained model & Parameters & ROUGE1 & ROUGE-L & BLEU1 & BLEU4 & +2 & +1 & +0 & Score   \\
        \midrule
        \multicolumn{1}{c}{\multirow{3}{*}{Fine-tuning}} 
        & BART & 220M & \textbf{31.64} & \textbf{29.95} & \textbf{35.02} & \textbf{12.46} & 3.2\% & \textbf{81.4\%} & \textbf{15.4\%} &  0.878 \\
        & EVA2.0 & 300M & 25.18 & 23.29 & 31.60 & 8.25 & 1.5\% & 67.6\% & 30.9\% &  0.706 \\
        & CDialGPT & 104M & 18.98 & 17.42 & 28.54 & 7.42 & 2.9\% & 38.5\% & 58.6\% & 0.443 \\
        
        \midrule
        \midrule
        \multicolumn{1}{c}{\multirow{2}{*}{1-Shot}} 
        & GLM & 10B & 18.44 & 16.99 & 29.48 & 7.26 & 12.6\% & 61.7\% & 25.7\% &  0.868 \\
        
        & GPT3 & 175B & 13.87 & 12.10 & 23.98 & 5.84 & 11.4\% & 56.3\% & 32.3\% & 0.791 \\
        \midrule
        \multicolumn{1}{c}{\multirow{2}{*}{8-Shot}} 
        & GLM & 10B & 20.72 & 19.22 & 28.78 & 7.70 & \textbf{14.9\%} & 65.0\% & 20.1\% & \textbf{0.949}  \\
        
        & GPT3 & 175B & 18.87 & 16.80 & 29.05 & 7.69 & 10.8\% & 66.3\% & 22.8\% & 0.880 \\
	\bottomrule
	\end{tabular}
        }
	\caption{\label{tab:transfer} Automatic and human evaluations from different pre-trained generative models. The 2/1/0 score schema is elaborated in Appendix \ref{subsec:metrics}. Score is the average score.}
\end{table*}

\myparagraph{Influence of Text Profiles} For the extraction of our text profiles, we empirically analyze the effect of different extraction methods as illustrated in Table \ref{tab:selection}. The \textit{random from user} means we randomly select sentences by the streamer as his or her text profiles, and \textit{random from dataset} refers to randomly selected in the whole dataset. The \textit{Length} represents the maximum truncation length for all concatenated text profiles. We can see that the rules and classifier both improve the model performance, indicating rules can filter the noisy sentences to some extent and persona definition in \texttt{DuLemon} is effective for training a classifier to further refine text profiles. Besides, the increase in persona sentence length will also enrich persona profiles and improve the results.

\subsection{Results of Addressee Recognition}
Previous works ~\citep{gu-etal-2021-mpc,le-etal-2019-speaking} adopt BERT to classify the relationship between the streamer response and multiple user comments, and we adopt a similar approach with a step further to explore the benefits of persona profiles.
TwinBERT, compared with BERT, is utilized to study the difference between single-stream and double-stream architecture, and CoBERT is for investigating the influence of our collected persona profiles.

Table \ref{tab:addressee} presents the results of addressee recognition. It shows that single-stream BERT outperforms double-stream TwinBERT. The reason is that by feeding the concatenation of the context and the response into a unified BERT, the interaction and aggregation operations can be performed through the attention mechanism sufficiently. Besides, CoBERT retrieves a better performance than TwinBERT, demonstrating our persona profiles are also beneficial to addressee recognition.

\section{Transfer Learning}
\begin{figure}[!t]
	\centering
	\includegraphics[width=0.45\textwidth]{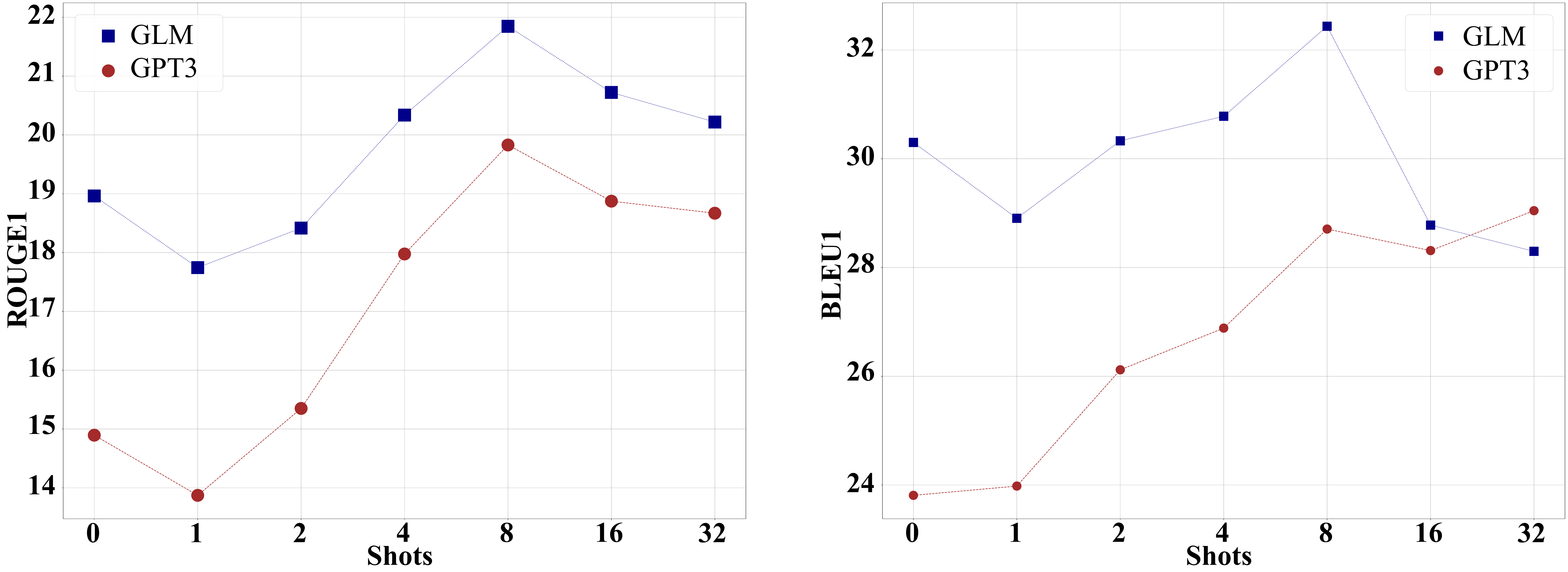}
	\caption{In-context learning results of GLM and GPT3 on different shots.}
	\label{fig:incontext}
\end{figure}

To further investigate the performance of the pre-trained dialogue model on our LiveChat, we fine-tune BART, Chinese CDialGPT, and EVA2.0 to study whether pre-trained dialogue corpora can contribute to the learning of our case. The latter two are trained on dialogue data from text-based social media.
Furthermore, we conduct in-context learning on GLM and GPT3 to explore the few-shot transferability of large language models (LLMs) on this video-sourced dataset.
The data utilized in Table \ref{tab:transfer} and Figure \ref{fig:incontext} are dissimilar, and the details of the training data as well as our in-context templates are expounded upon in Appendix \ref{appendix:Generation}.

Table \ref{tab:transfer} shows the results.
First, the performance of BART is better than EVA2.0 and Chinese DialGPT. It confirms that the domain of our LiveChat is far away from the domains of those dialogue datasets utilized in existing pre-trained dialogue models.
Therefore, it is challenging work to directly transfer from models trained on other dialogue domains.
LLMs, nevertheless, offer a solution to this problem due to their great ability to generalization. 
Although the automatic evaluation results of fine-tuned models are better than LLMs by the reason that fine-tuning enables the models to learn the intrinsic distribution of LiveChat.
We discover that the percentage of score 2 in human evaluation results of LLMs is dramatically larger than fine-tuned ones, which means better performance in terms of rich informativeness.
We attribute this to the massive knowledge contained in LLMs and the few-shot demonstrations to elicit such knowledge. Yet despite this, we see a performance gap in score 1 with BART, which indicates a large room to increase contextual coherence through ways like parameters-efficient domain adaptation of LLMs to LiveChat, simultaneously maintaining their original powerful capabilities.  

As a supplement, we also have performed a series of experiments of in-context learning on different shots to study the influence of demonstrations.
The ROUGE1 and BLEU1 results are depicted in Figure \ref{fig:incontext}. The performances keep growing as the shots gradually increase. However, when the number of demonstrations exceeds 8 shots, the performances of the LLMs slightly decrease due to the random manual selection of demonstrations. 





%% file: sections/6conclusion.tex
\section{Conclusion}

In this paper, we propose LiveChat, a Chinese video-sourced and personalized dialogue dataset from the live streaming domain with detailed persona profiles.
It maintains the largest average sessions per persona and is also the largest MPC dataset for addressee recognition since live streaming is a natural MPC scenario. This is achieved owing to the reply-to-whom matching method that enables automatically extracting dialogue sessions from live videos, while most video extraction methods can not.
Experimental results on two benchmark tasks show that the selected persona profiles and the larger number of average sessions per persona are advantageous in learning the speaker's personalized response and addressee decision. 
In addition, the comparisons between BART with other pre-trained dialogue models and LLMs have unveiled the distinctiveness of this video-sourced dialogue domain and we expect further research on parameters-efficient transfer learning of LLMs for LiveChat.


%% file: sections/7limitations.tex
\section*{Limitations}
There exist some limitations in our work. LiveChat is a Chinese-originated dataset involving unique cultures and abundant replying styles. However, this intensifies the difficulty of fully understanding the content of this dataset. Fortunately, the same data construction pipeline can be applied to streaming platforms of other languages, like TikTok. And currently, our LiveChat is only sourced from 351 streamers on Douyin, not sufficient to train a general chatbot. We believe that LiveChat helps get one's foot in the door to the wonderful and diversified live scenarios and a dialogue model pre-trained on the considerable amount of video-sourced dialogue data among cross-platforms is promising.
Besides, LiveChat contains some noisy spoken language segments that are not easy to read after transcribing from the ASR tool. The upper bound data quality is limited by such third-party tools. The future work to concatenate such text segments to restore the content of the original expression by streamers is highly anticipated. As for the dialogue-matching method, we simply implement a combination of BOW and BERT for semantic matching, which needs further optimization.

Other limitations from the training perspective can also be highlighted. For example, contextual background information is not considered in our modeling. That includes history dialogues in multi-turn settings and information from other modalities, like the streamer eating in front of the camera. In addition, we have not explored enough of our annotated basic profiles. In our primary experiments, we found that directly adding basic information such as age, gender, location, and other room information has limited influence on the model performance. We account for the fact that these basic profiles have limited connections with reply styles and contents in LiveChat. Also, note that we remove the repetition part of a streamer's response before training, while it is useful to maintain this pattern in practical application.

%% file: sections/8ethicalconsideration.tex
\section*{Ethical Consideration}
\label{sec:ethical} 
This work presents LiveChat, a free and open Chinese dataset for the research community to study personalized open-domain dialogue generation and addressee recognition. Our dataset contains well-processed dialogues, and annotations (basic profiles and text profiles). 

\myparagraph{Data Privacy} The original live-streaming clips and streamers' profiles of LiveChat are collected from Douyin, one of the largest Chinese live-broadcasting platforms. Similar to previous dialogue data from Reddit~\citep{mazare-etal-2018-training} and Weibo~\citep{pchatbot},
LiveChat is an open-domain dialogue dataset that crossover multiple topics and users. 
Since all streamers must comply with platform rules during their online live streaming under the strict supervision of the Chinese government, their topics do not contain any pornographic, violent, reactionary, or discriminatory statements. Besides, due to the property of streaming, historically broadcast videos are no longer available when finished. Therefore it is not traceable from LiveChat to the identity of real streamers.
Moreover, we clean the raw data with transformation, anonymization, and deletion to ensure there is no disclosure of private information and the identity of the streamers or audiences can not be inferred from it. 
Thus, all the collected data (including persona profiles) is publicly available and does not contain any private information of streamers and audiences, such as emails, phone numbers, and real user names. Although we collect the Age and Location information,
in our basic profile, the Age is expressed as an interval range that doesn't represent the real age of the streamers,
and the Location only contains the province's information. Besides, all the attributes of our basic profiles are re-indexed as numbers in the final released dataset.
Thus, both our raw data and persona profiles do not create additional ethical risks. Moreover, we are sure that all the collected data is consistent with the platform usage rules and protocols. 
LiveChat will only be allowed to be used for academic research. At last, our construction of LiveChat was approved by an internal review board (IRB).

\myparagraph{Annotators} In terms of basic profile annotation and manual evaluation, all the annotators are Chinese undergraduates specifically responsible for annotation work in our institution. They are informed of the ongoing research and well known the way the curated data will be used. All the annotated information and evaluation results do not contain any private information.



%% file: sections/10Acknowledgements.tex

%% file: sections/9appendix.tex
\label{sec:appendix}


\section{Dataset Construction Details}
\label{sec:data const}
Our constructed dataset are composed of 1332073 dialogues, and each dialogue consists of one streamer response and several audience comments. The overall statistics of the LiveChat and raw data are illustrated in Table \ref{tab:Statistic}.

\myparagraph{Details of Automatic Speech Recognition}\label{ASR} Our HuoShan ASR tool is from Chinese company ByteDance. The ASR is pretrained on a large entertainment dataset that includes domains such as fashion, food, games, and singing. After testing on a 64k Chinese video-based recognition dataset from various domains, the ASR achieved a Character Error Rate (CER) of 3.17\%.

\begin{figure}[htbp]
	\centering
	\includegraphics[width=0.46\textwidth]{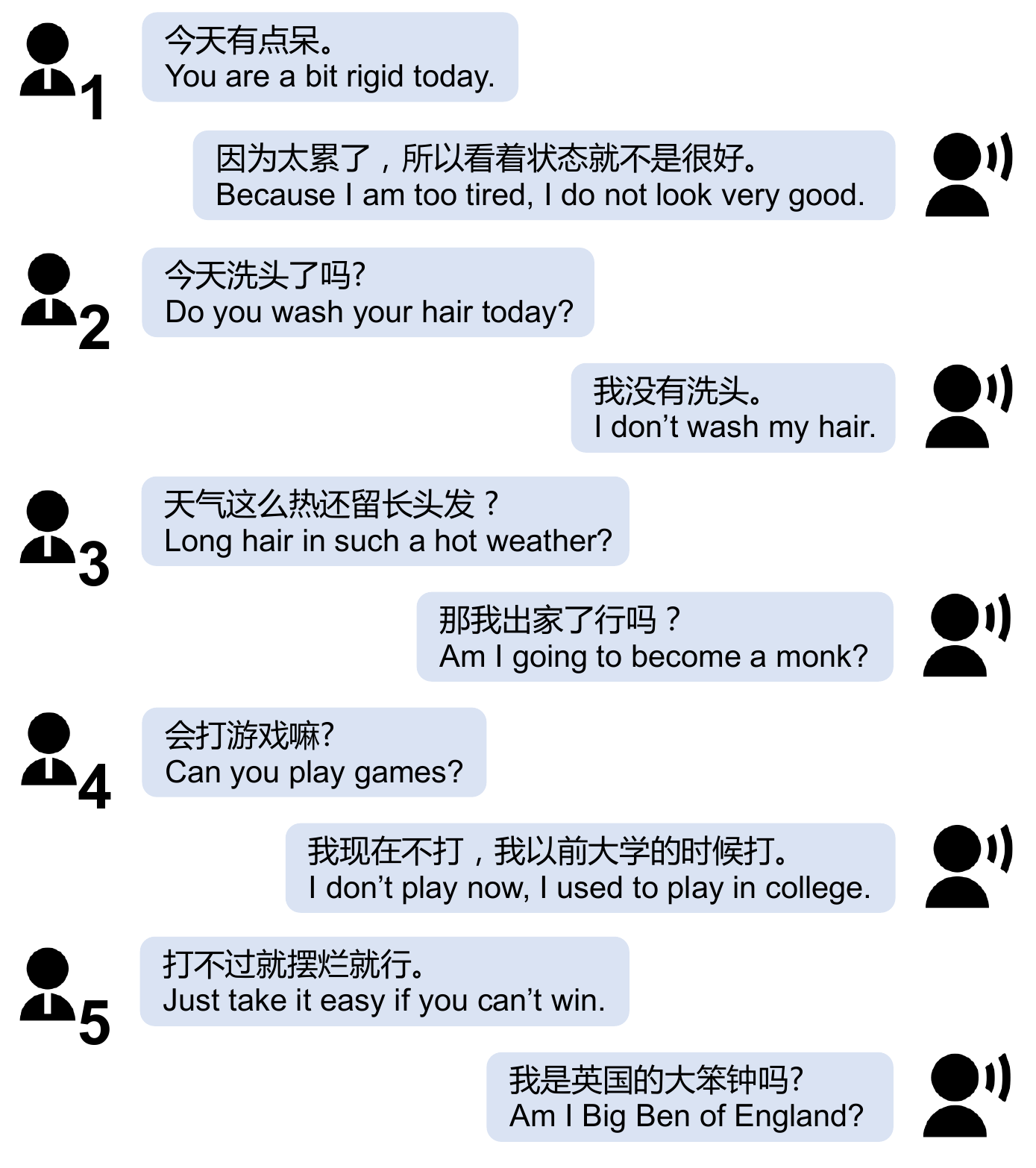}
	\caption{A conversation between one streamer and several audiences in LiveChat.}
	\label{fig:DialogueCases}
\end{figure}

\myparagraph{Dialogue samples in response modeling} In response modeling, we select all the matched dialogue pairs from our raw conversation dataset. Several constructed dialogue cases are shown in Figure \ref{fig:DialogueCases}. Each audience comment is associated with a streamer response. During our retrieval-based response modeling experiments, given an audience comment, all the responses in one batch are negative responses. 

\begin{figure}[htbp]
	\centering
	\includegraphics[width=0.46\textwidth]{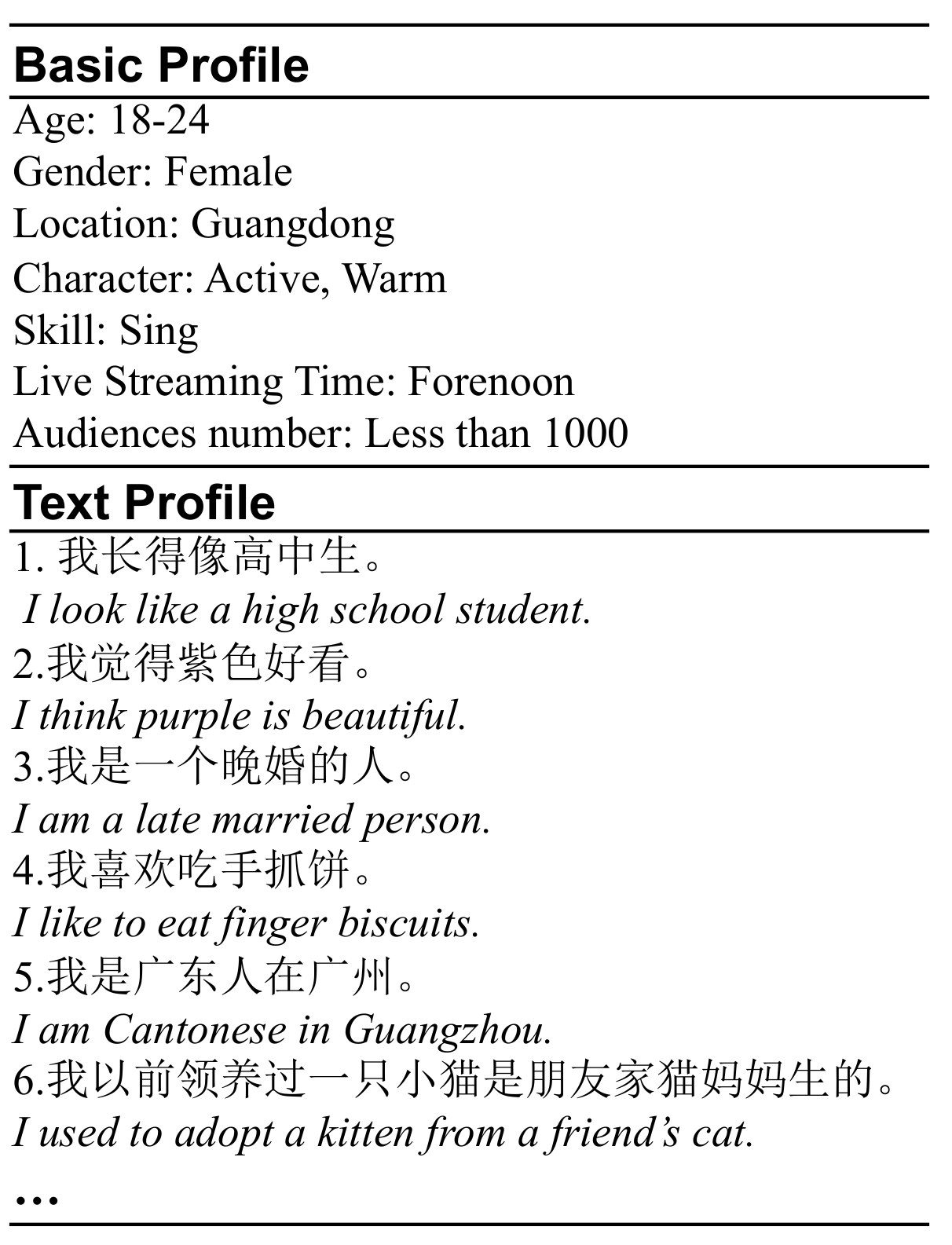}
	\caption{The annotated basic profile and collected text profile of one streamer. Note that in the final released dataset, all basic profiles are re-indexed as numbers for privacy concerns.}
	\label{fig:profiles}
\end{figure}

\begin{figure}[htbp]
	\centering
	\includegraphics[width=0.46\textwidth]{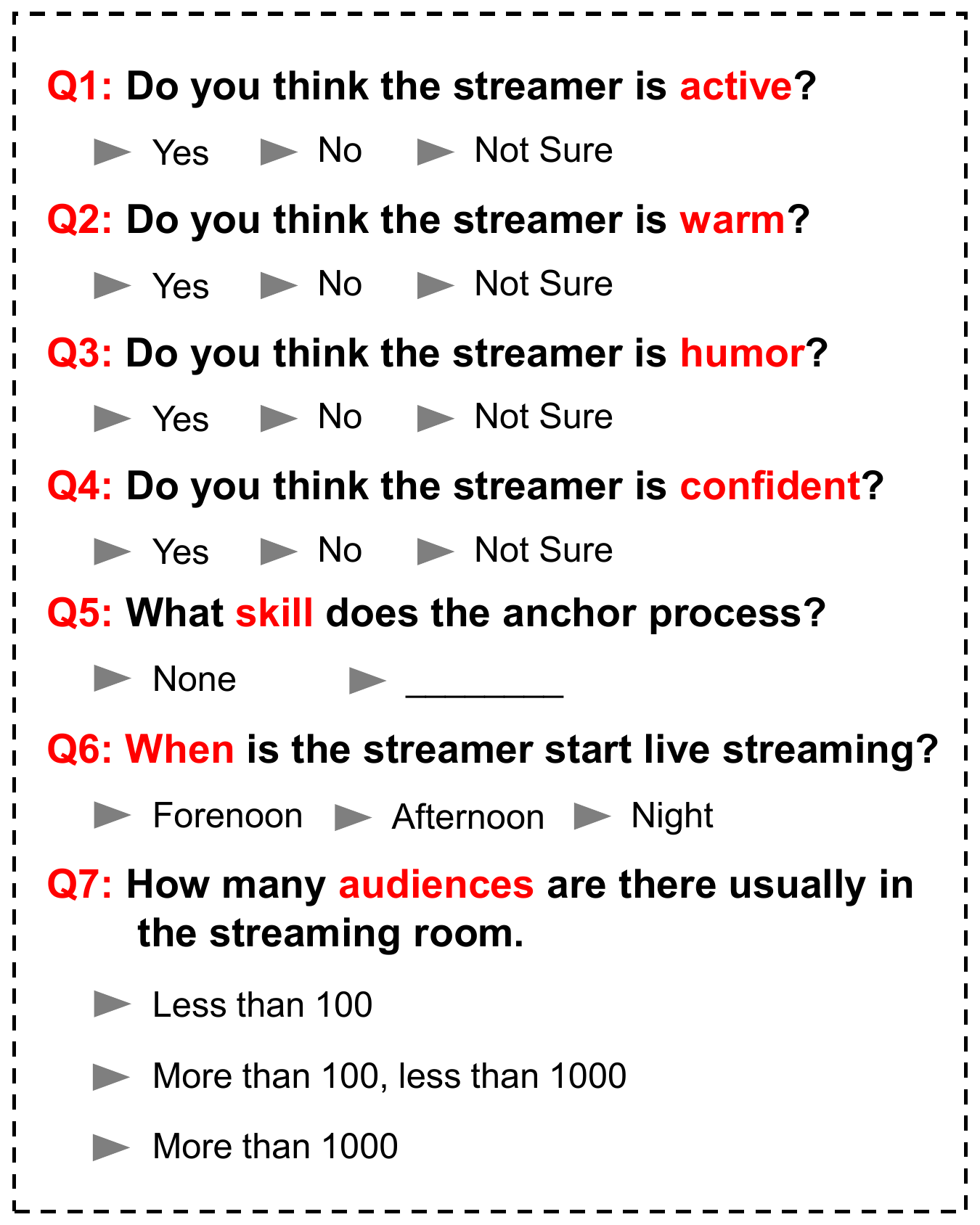}
	\caption{Annotation User Interface.}
	\label{fig:Interface}
\end{figure}

\myparagraph{Persona Annotations} Our persona annotations include the basic profile and text profile, and a persona profile sample of one streamer is shown in Figure \ref{fig:profiles}. Text profiles are collected from the history posts and dialogues based on the rules and a persona classifier, and basic profiles are collected and annotated by crowdworkers who are native Chinese speakers and familiar with live streaming. Apart from the basic information on the streamer's homepage, the crowdworkers are required to label some extra information that may have an influence on the streamer's speaking style. We present our annotation interface in Figure \ref{fig:Interface}. For each streamer, the annotator is required to answer these questions based on the provided live streaming videos. 

\begin{figure}[htbp]
	\centering
	\includegraphics[width=0.48\textwidth]{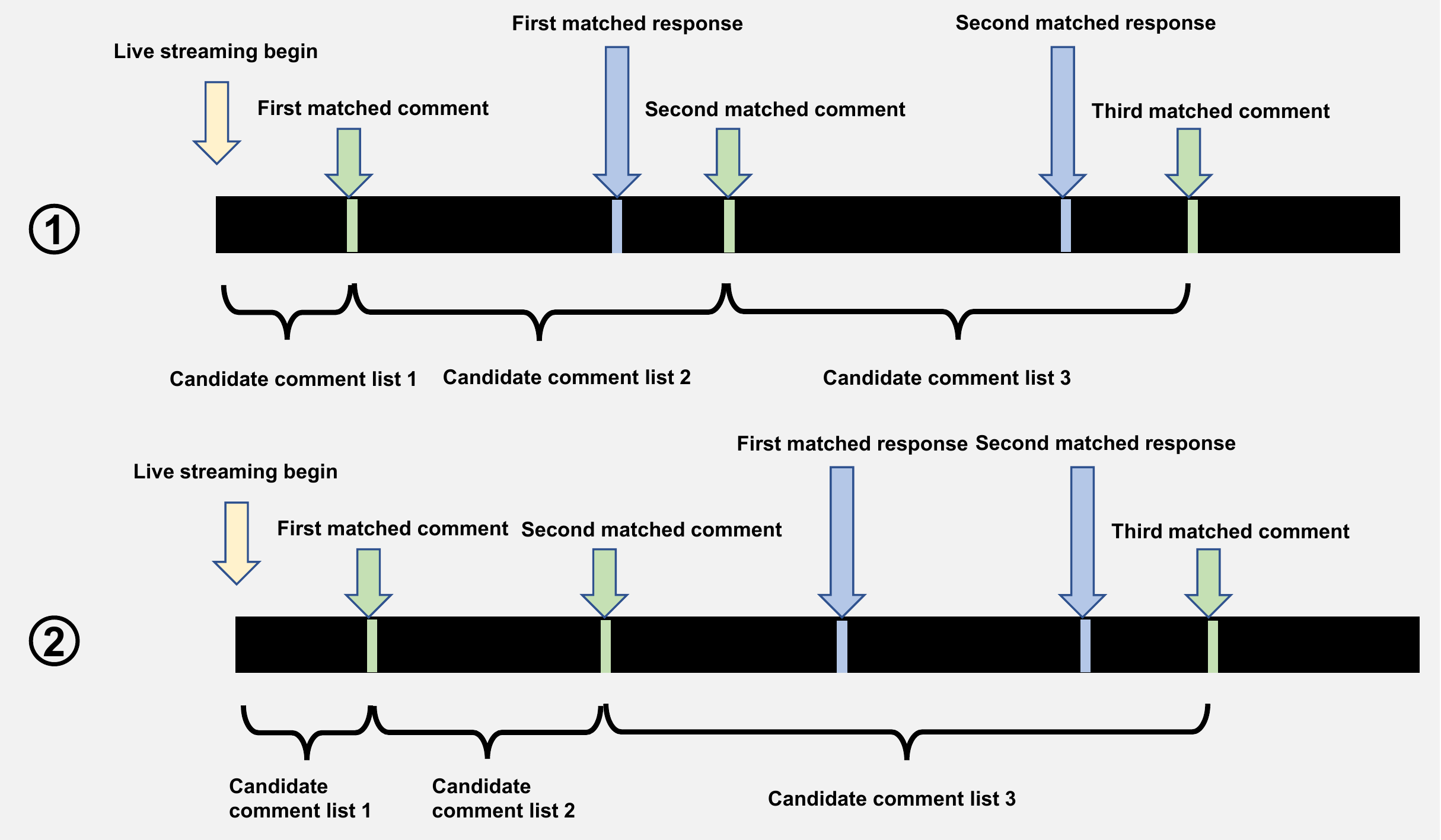}
	\caption{Segmentation for candidates of comments.}
	\label{fig:Selection}
\end{figure}

\myparagraph{Selection of candidate audiences} \label{appendix:Selection} A streamer in LiveChat will respond to one audience selectively, and the segmentation of all audience comments is shown in Figure \ref{fig:Selection}. We noted the timestamp of the matched comments and responses among all the comments. The comments between matched ($i-1$)-th comment and $i$-th comment are the candidate comments of the streamer's $i$-th response. In addressee recognition, the streamer aims to retrieve which comment among these candidates to respond to.

\begin{table}
	\centering
	\footnotesize
	\setlength{\tabcolsep}{1.5mm}
	\begin{tabular}{l|c}
	\toprule
        Category & Size \\ 
        \midrule
       Raw Audiences Comments & 13,745,577 \\
       Raw Total Video Num. & 182,943 \\
       Raw Total Videos Hours & 30,248 \\
       Raw Streamer Sentences & 35,475,979 \\
       Dialogues &  1,332,073  \\
       Utterances &  9,416,573 \\
       Streamer Num. &  351 \\ 
       Audience Num. &  1,058,595 \\
       Avg. Sessions per Streamer & 3,795 \\
       Avg. Length of Utterances &  10.22 \\
       Avg. Sentences of Text Profiles &  69 \\

	\bottomrule
	\end{tabular}
	\caption{\label{tab:Statistic} Statistic of LiveChat.}
\end{table}


\begin{figure}[htbp]
	\centering
	\includegraphics[width=0.48\textwidth]{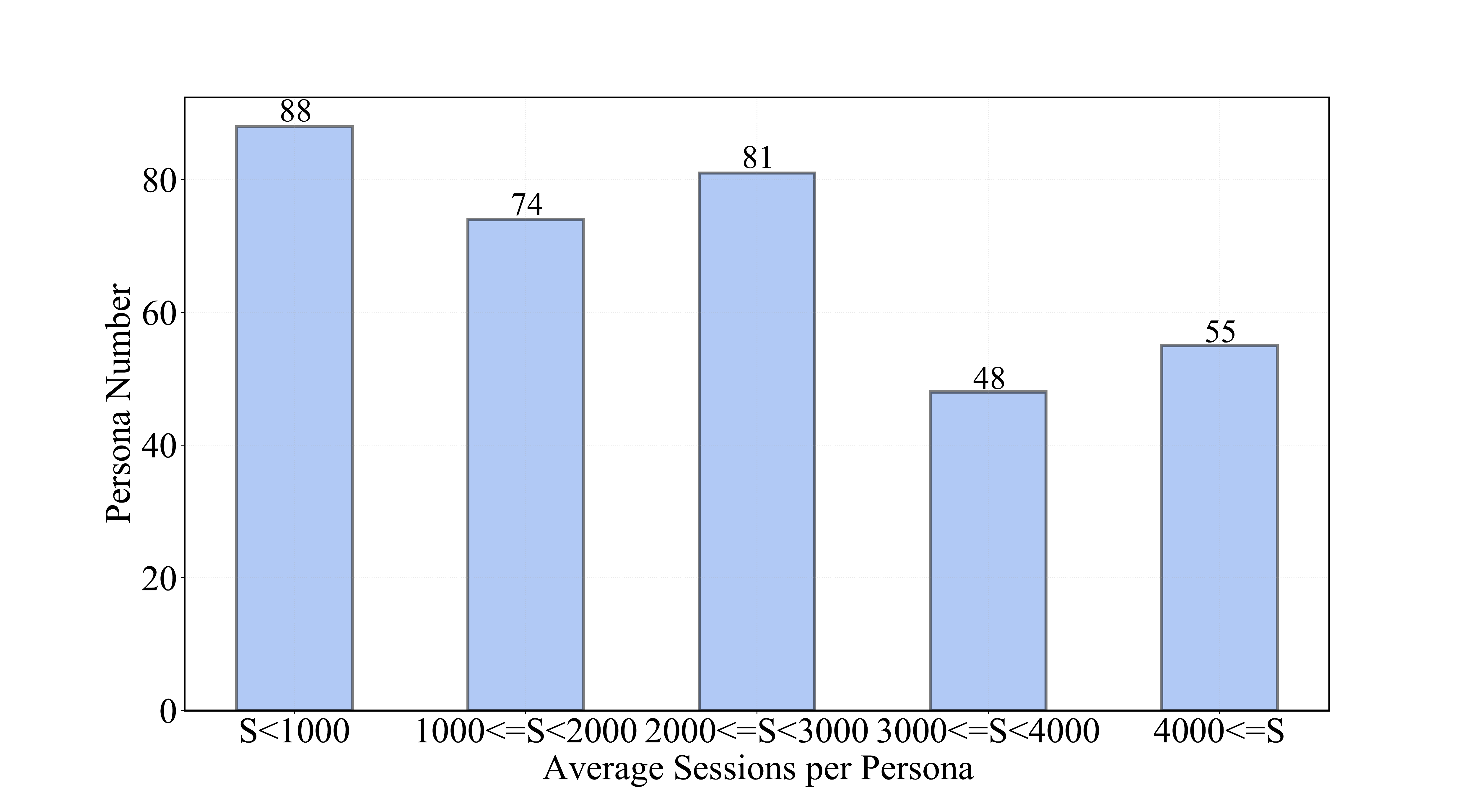}
	\caption{Session length distribution in LiveChat.}
	\label{fig:session}
\end{figure}

\section{Training and Evaluation Details} \label{sec:train details} 
\subsection{Training Details}

\myparagraph{Retrieval-based models} Figure \ref{fig:session} provides the distribution of session length for each persona. There exist some persona IDs without enough sessions, thus we filter those IDs with more than 2400 sessions to study the influence of the average session number and persona profiles in a more clear setting. In this way, we retrieve 150 persona IDs in total. During our training process, we use 400k dialogues for training and 10k dialogues for testing in all retrieval-based dialogue experiments if there is no declaration before. The batch size is set to 24, which also means the size of the dynamic searching library of response modeling is 24. 

In addressee recognition, the number of candidate comments ranges from one to hundreds. Thus, we process each session into one response and 10 candidate comments. If comments are too many, we select the last 10 comments, where the final sentence is the corresponding comment. And if the number of comments in one session is less than 10, we add comments in the front sessions to keep the total comment number to 10 in each session. The batch size we set here is also 24. 

During training, we set the max input length and output length as 64, the max text profiles length as 512, and the epoch number and learning rate are set to 30 and 1e-5. All the experiments in the above two dialogue tasks are conducted on Nvidia Tesla V100s.

\myparagraph{Generation-based models}\label{appendix:Generation} During the process of fine-tuning the pre-trained language models, we keep the most original experimental settings from their initial training parameters, and the utilized GPT3 version is text-davinci-002. In Table \ref{tab:transfer}, the training dataset for fine-tuning is 400k, and the test dataset is 10k. Due to the cost of the GPT3 API, we only evaluate 1k samples for each experiment of GPT3 in Figure \ref{fig:incontext}. In order to keep in line with GPT3, all data utilized in GLM is the same as GPT3. Thus, the results in Table \ref{tab:transfer} are inconsistent with those in Figure \ref{fig:incontext}.

As for the in-context learning of GLM and GPT3, the template of n-shots is formulated as "我是一名线上直播间的主播，爱好是唱歌、与粉丝聊天等。以下是我在直播间和粉丝的互动。粉丝说：\textsc{[Context-1]}。我说：\textsc{[Response-1]}。...粉丝说：\textsc{[Context-n]}。我说：\textsc{[Response-n]}。以下是另一段我在直播间和粉丝的互动。粉丝说：\textsc{[Context-test]}。 我说：\textsc{[Response-test]}" ("I am a streamer of an online live room, hobbies are singing, chatting with fans and so on. Followings are my interactions with fans in the live room. One fan says: \textsc{[Context-1]} I say: \textsc{[Response-1]} ... One fan says: \textsc{[Context-n]} I say: \textsc{[Response-n]}. Here is another interaction I have with my fans in the live room. One fan says: \textsc{[Context-test]} I say：\textsc{[Response-test]}). 

The \textsc{[Context-k]} and \textsc{[Response-k]} ($0 < k <= n$) is the n-shot cases provided for LLMs. The \textsc{[Context-Test]} and \textsc{[Response-Test]} are the two utterances of one test dialogue pair, where the LLMs are required to return the \textsc{[Response-Test]}.

\subsection{Metrics}
\label{subsec:metrics}
\myparagraph{Retrival-based} \textbf{Recall@k} is a commonly used metric for evaluating whether the correct response exists among the top k candidates out of all the candidate responses. \textbf{MRR} (Mean Reciprocal Rank) is a statistic measure for evaluating any process that produces a list of possible responses to a sample of queries and is formulated as the average of the reciprocal ranks of results.



\myparagraph{Generation-based} \textbf{BLEU-n} measures the ratios of the co-occurrences of n-grams between the generated and real text. \textbf{ROUGE-n} measures the text quality by counting the overlapping n-grams between the generated and real text, and \textbf{ROUGE-L} means leveraging the longest common subsequence. 


\myparagraph{Human Evaluation} We employ crowd workers to evaluate the responses generated by different models, and 1000 samples for each model. Our evaluation schema scores each sentence according to the following rules, inspired by~\citet{CDialoGPT}:
\begin{enumerate}
    \item \textbf{Relevance} If a fluent response is logically consistent and relevant to the content of the comment, it will get 1. Otherwise, it will get 0.
    \item \textbf{Informativeness} If a response has achieved the requirements of 1 score and is additionally rich in content, it will get 2.
\end{enumerate}

\section{Case Study}

\begin{table*}
	\centering
	\footnotesize
	\setlength{\tabcolsep}{1.5mm}
	\begin{tabular}{l|l|l}
	\toprule
        Context & \makecell[c]{搞笑女来了。 \\ The comedian girl has arrived.} & \makecell[c]{我一个月的工资都刷给你了。 \\ I have spent all of my salaries for the month on you.} \\
        \midrule
        \midrule
        BART & \makecell[c]{我不是搞笑女我是颜值主播。 \\ I am not a comedian, \\ I am a good-looking anchor.} & \makecell[c]{你一个月工资多少啊? \\ How much do you make a month?} \\
        
        EVA2.0 & \makecell[c]{搞笑女孩我来了。 \\ The funny girl is here.} & \makecell[c]{你这个工资是不是很高啊。 \\ Your salary is not very high.} \\
        C-DialGPT & \makecell[c]{可是我很少活的很有感情。 \\ But I rarely live with much feeling.} & \makecell[c]{你为什么可以刷到9,000？ \\ Why can you achieve 9,000?} \\

        GLM (1-shot) & \makecell[c]{对呀,我是搞笑女。 \\ Yeah, I am a comedian girl. } & \makecell[c]{你刷得有点多哦。 \\ You are doing a little too much.}\\
        GPT3 (1-shot) & \makecell[c]{是谁来了？ \\ Who is there?} & \makecell[c]{你辛苦了，我会努力争取更多粉丝。 \\ Thank you. I'll try to get more fans.} \\
        GLM (8-shot) & \makecell[c]{搞笑女来了,快进来,快进来。 \\ The comedian girl is here, \\ come in quickly, come in quickly. } & \makecell[c]{刷?你不许刷。 \\ Spend? You are not allowed to do that.}\\
        GPT3 (8-shot) & \makecell[c]{你要搞笑，我可不搞笑. \\ You want to be funny, \\ I do not want to be funny.} & \makecell[c]{你一个月工资都刷给我了？你不用上班了？ \\ You spent your entire month's salary on me?  \\ You don't have to work anymore?} \\
        \midrule
        \midrule
        Expected Response & \makecell[c]{不不皮了。 \\ No, do not joke.} & \makecell[c]{是听到我们的号召了吗？ \\ Is that for hearing the call of us?} \\
	\bottomrule
	\end{tabular}
	\caption{\label{tab:casestudy} Generated responses by different pre-trained language models.}
\end{table*}

\begin{table*}
	\centering
	\footnotesize
	\setlength{\tabcolsep}{1.5mm}
	\begin{tabular}{l|l|l}
	\toprule
        Context & \makecell[c]{跳个舞蹈吧。 \\ Have a dance.} & \makecell[c]{走错直播间了？ \\ In the wrong live streaming room?} \\
        \midrule
        \midrule
        BART & \makecell[c]{你们想看什么舞蹈？ \\ What dance do you want to see?} & \makecell[c]{没有走错直播间。 \\ Haven't gone to the wrong live streaming room.} \\
        
        EVA2.0 & \makecell[c]{你给我跳个跳舞吧。 \\ I want you to dance for me.} & \makecell[c]{没有啊,他们都说我在直播。 \\ No, they all say that I am live streaming.} \\
        C-DialGPT & \makecell[c]{感谢落落谢谢无心球。 \\ Thank Luoluo, thank Wuxin Ball.} & \makecell[c]{你好，仙人掌花开放26度。 \\ Hello, the cactus is blooming at 26 degrees.} \\

        GLM (1-shot) & \makecell[c]{好呀好呀,不过呢,现在得先唱歌。 \\ Great, however, now we have to sing first. } & \makecell[c]{没有。 \\ No.}\\
        GPT3 (1-shot) & \makecell[c]{不行，我不会跳舞。 \\No, I don't know how to dance.} & \makecell[c]{你没有走错，这里是我的直播间， \\ 你回去收看我的直播吧。 \\ You did not go wrong, this is my live streaming room, \\ please go back and watch my live streaming.} \\

        GLM (8-shot) & \makecell[c]{我现在心情不是很好。 \\ I am not feeling well now. } & \makecell[c]{走错直播间了吗? \\ Have you gone into the wrong live streaming room?}\\
        GPT3 (8-shot) & \makecell[c]{跳个舞蹈要不然？ \\Why not have a dance?} & \makecell[c]{怎么你问这个，走错我的直播间了吗？ \\ Why are you asking this, did you accidentally \\ go to the wrong live streaming room of mine?} \\        
        \midrule
        \midrule
        Expected Response & \makecell[c]{我不跳要不你给我跳一个看看。 \\ I don't dance and you can dance for me.} & \makecell[c]{你没有走错，这是在下。 \\ You haven't taken a wrong turn, this is me.} \\
	\bottomrule
	\end{tabular}
	\caption{\label{tab:casestudy2} Generated responses by different pre-trained language models.}
\end{table*}

To concretely demonstrate response ability in generative-based  models, we further provide several sample cases (between BART, C-DialGPT, EVA2.0, GLM, and GPT3) in Table \ref{tab:casestudy} and Table \ref{tab:casestudy2}. BART, C-DialGPT, and EVA2.0 are finetuned on our LiveChat. LLMs present the results of 1-shot and 8-shot in-context learning.

\myparagraph{Relevance} Seen in these cases, all the responses generated by BART have strong connections with input contexts. Compared to EVA and C-DialGPT, BART maintains a higher relevance. For example, in Case 2-"\texttt{走错直播间了？(In the wrong live streaming room?)}"  from Table \ref{tab:casestudy2}, we can find the response of C-DialGPT is not logically relevant to the comment, and the response of EVA2.0 is also not reasonable. 

\myparagraph{Informativeness} Pre-trained models generally contain knowledge inside themselves. We can see that LLMs reply with more informative content in some cases, which means the richness and abundant knowledge of LLMs will be leveraged in dialogue generation.